\pdfoutput=1
\PassOptionsToPackage{table}{xcolor}
\documentclass{article}

\usepackage[preprint]{neurips_2026}

\usepackage[utf8]{inputenc}
\usepackage[T1]{fontenc}
\usepackage{hyperref}
\usepackage{url}
\usepackage{booktabs}
\usepackage{amsfonts}
\usepackage{amsthm}
\usepackage{amsmath,amssymb,mathtools}
\usepackage{nicefrac}
\usepackage{microtype}
\usepackage{inconsolata}
\usepackage{latexsym}
\usepackage{graphicx}
\usepackage{subcaption}
\usepackage[most]{tcolorbox}
\usepackage{enumitem}
\usepackage{xcolor}
\usepackage{wrapfig}
\usepackage{amssymb}
\usepackage[normalem]{ulem}

\DeclareMathOperator{\mean}{mean}
\DeclareMathOperator{\std}{std}

\newtcolorbox{takeawaybox}[1][]{
  colback=green!5!white,
  colframe=green!30!black,
  title=\textbf{TAKEAWAY},
  fonttitle=\bfseries,
  arc=3mm,
  left=2.5mm,
  #1
}
\newtcolorbox{questionbox}[1][]{
  enhanced,
  colback=blue!3!white,
  colframe=blue!45!black,
  boxrule=0.5pt,
  arc=2mm,
  left=2.5mm,
  right=2.5mm,
  top=1.8mm,
  bottom=1.8mm,
  drop shadow,
  #1
}

\title{Asymmetric Advantage Modulation Calibrates Entropy Dynamics in RLVR}
\author{
  Hengrui Gu\textsuperscript{1} \quad
  Xiaotian Han\textsuperscript{2} \quad
  Yujing Bian\textsuperscript{1} \quad
  Feiyi Wang\textsuperscript{3} \quad
  Kaixiong Zhou\textsuperscript{1} \\[6pt]
  \textsuperscript{1}North Carolina State University \quad
  \textsuperscript{2}Case Western Reserve University \quad
  \textsuperscript{3}Oak Ridge National Laboratory \\[3pt]
    \texttt{hgu6@ncsu.edu} \quad
    \texttt{xhan@case.edu} \quad
    \texttt{ybian3@ncsu.edu} \quad
    \texttt{kzhou22@ncsu.edu}
}

\begin{document}
\maketitle
\begin{abstract}
Reinforcement learning with verifiable rewards (RLVR) has substantially improved the reasoning ability of large language models (LLMs), but it often suffers from \textit{restricted exploration}, where the policy rapidly concentrates on a narrow set of solutions. A common remedy is entropy regularization, which attempts to preserve exploration by increasing policy entropy. However, for LLM-RL, this intervention is highly sensitive to its coefficient, can introduce semantically weak uncertainty, and often yields limited accuracy gains. This motivates a more precise question: which entropy helps reasoning, and which entropy should be reduced? To study this, we parameterize the advantage estimator in Group Relative Policy Optimization (GRPO) into positive and negative outcome-conditioned channels and analyze their entropy dynamics. Our results show that positive-channel modulation raises \textit{productive entropy} associated with successful reasoning trajectories, while negative-channel modulation removes \textit{noisy entropy} associated with failed rollouts and reduces interference with correct paths. Guided by this channel-wise view, we propose \textbf{AsymGRPO}, which decouples the modulation strengths of positive and negative advantages. This enables flexible control over how the model updates across prompt difficulty levels, allowing stronger reinforcement of rare successes on harder prompts or stronger suppression of residual failures on easier prompts without forcing the two channels to share the same modulation strength. Experiments on five mathematical reasoning benchmarks show that AsymGRPO outperforms strong RLVR baselines, with consistent gains across model backbones.

\end{abstract}

\section{Introduction}
Reinforcement learning with verifiable rewards (RLVR) has emerged as a central post-training paradigm for improving the reasoning ability of large language models (LLMs)~\cite{zhang2025survey,lambert2024tulu,wen2025reinforcement,mroueh2025reinforcement,wen2025reinforcement,lv2025towards}. Instead of relying on learned reward models, RLVR uses automated verifiers to provide programmatic correctness feedback for complete model outputs. This verification signal alleviates reward-model overoptimization (``reward hacking'')~\cite{miao2024inform,gao2023scaling} and enables verification-guided search over reasoning trajectories~\cite{setlur2024rewarding,wang2025beyond}, leading to substantial gains on mathematical and coding tasks~\cite{gehring2024rlef,setlur2024rl}.

Despite these gains, RLVR faces a fundamental limitation known as \textit{restricted exploration}, which often appears as \textit{entropy collapse}~\citep{cui2025entropy,yu2025dapo,yue2025does}. Early in training, the policy can become overconfident in a narrow set of solutions, causing policy entropy to drop sharply and reducing the chance of discovering alternative reasoning paths. This collapse is not merely a change in a diagnostic metric; it indicates that potentially useful solution strategies may be removed before they are sufficiently explored, which can lead to premature performance saturation.


The standard response is to add entropy regularization to the training objective, with the expectation that higher policy entropy will preserve exploration~\citep{wang2025reinforcement,he2025skywork}. However, recent studies show that entropy regularization is less reliable for LLM-RL than in conventional RL~\cite{haarnoja2018soft,ppo}. It is highly sensitive to its coefficient, can cause entropy explosion toward near-uniform and semantically weak policies, and often yields only marginal performance gains~\citep{sirenentropy,entropycontrol,he2025skywork}. These failures suggest that the core issue is not simply whether entropy is large or small, but whether the entropy being preserved is useful for reasoning.


\begin{questionbox}
\centering
\textbf{Question:} \textit{Does simply increasing policy entropy actually produce better exploration in RLVR, or must training identify which entropy helps reasoning and which entropy hurts it?}
\end{questionbox}

To answer this question, we use the advantage estimator from Group Relative Policy Optimization (GRPO)~\citep{shao2024deepseekmath} as our analytical probe, because it allows entropy changes to be examined separately for successful and failed outcomes. For each prompt, GRPO samples a group of rollouts and computes the group accuracy \(p\). We parameterize its advantage estimator into two outcome-conditioned channels: a positive rollout receives \(A_{\mathrm{pos}}^{(\beta)}(p)=((1-p)/p)^{\beta}\), while a negative rollout receives \(A_{\mathrm{neg}}^{(\beta)}(p)=-(p/(1-p))^{\beta}\). Standard GRPO applies the same modulation strength to both channels, corresponding to \(\beta=0.5\). Using this parametric form to analyze entropy dynamics and reasoning accuracy, we find two channel-wise effects:
\begin{itemize}[leftmargin=*,topsep=2pt,itemsep=1pt]
    \item Positive-channel modulation raises policy entropy relative to vanilla RLVR by retaining probability mass on alternative successful reasoning trajectories. This higher entropy, which we call \textit{productive entropy}, is associated with improved reasoning accuracy, showing that entropy is beneficial when it expands useful solution paths.
    \item Negative-channel modulation reduces entropy associated with failed rollouts, mitigating interference with correct trajectories while also improving reasoning accuracy. The removed component, which we call \textit{noisy entropy}, shows that entropy reduction can also be beneficial when the suppressed uncertainty comes from unreliable outcomes.
\end{itemize}

Effective exploration in RLVR requires precise entropy refinement rather than blind entropy maximization. Guided by this insight, we introduce \textbf{Asymmetric Group-Relative Policy Optimization (AsymGRPO)}, a parametric generalization of GRPO that decouples the advantage modulation of successful and failed rollouts. For a prompt-level rollout group with accuracy \(p\), AsymGRPO assigns \(A_{\mathrm{pos}}^{(\beta_{\mathrm{pos}})}(p)=((1-p)/p)^{\beta_{\mathrm{pos}}}\) to positive rollouts and \(A_{\mathrm{neg}}^{(\beta_{\mathrm{neg}})}(p)=-(p/(1-p))^{\beta_{\mathrm{neg}}}\) to negative ones. By allowing \(\beta_{\mathrm{pos}}\) and \(\beta_{\mathrm{neg}}\) to vary independently, AsymGRPO can separately shift the concentration of positive and negative gradient pressure across prompt-difficulty levels, enabling stronger reinforcement of rare successes or stronger suppression of residual failures without forcing the two channels to share the same modulation strength. Experiments on five mathematical reasoning benchmarks demonstrate that AsymGRPO achieves highly
competitive performance compared to strong RLVR baselines, with consistent gains
across model backbones. We provide detailed related work on RLVR post-training,
entropy control, and advantage estimation for RLVR in
Appendix~\ref{app:related_work}.

\section{Decomposing Group-Relative Advantages into Opposing Channels}
\label{sec:entropy_mechanism}

This section reformulates group-relative advantages as a continuous family of
estimators and probes its entropy behavior analytically.
Subsection~\ref{subsec:parametric_grpo} introduces a $\beta$-parametrized
family that consumes two channels of advantages corresponding to 
successful and failed rollouts, respectively. Subsection~\ref{subsec:oppo_entropy} then uses
this family as an analytical probe, isolating each reward channel's entropy effect.
The resulting picture is clear: the positive and negative advantage channels exert
structurally opposite on policy entropy. This decomposition facilitates entropy-accuracy effect study and reward design.

\subsection{Unifying REINFORCE and GRPO via a Parametric Advantage Family}
\label{subsec:parametric_grpo}
RLVR is a policy-gradient framework in which an automated verifier assigns a binary reward to each sampled response. Given an LLM policy $\pi_\theta$, its objective is
\begin{equation}
\label{eq:rlvr}
\max_{\theta}\;\mathcal{J}_{\mathrm{RLVR}}(\theta)
=
\mathbb{E}_{x\sim\mathcal{D},\, y\sim\pi_{\theta}(\cdot\mid x)}
\big[\,r(x,y)\,\big],
\end{equation}
where $x$ is a prompt drawn from dataset $\mathcal{D}$, $y$ is a rollout produced by $\pi_\theta(\cdot\mid x)$, and $r(x,y)\in\{0,1\}$ is the verifiable correctness reward. Practical training typically replaces the raw reward $r(x,y)$ with a centered advantage estimate $A(x,y)$, obtained by subtracting a baseline, to reduce variance without changing the expected policy-gradient direction. We next compare REINFORCE, GRPO, and a continuous parametric family under common notation.

\paragraph{REINFORCE reference.}
The simplest instantiation of the advantage in Eq.~\eqref{eq:rlvr} is REINFORCE~\citep{williams1992simple}. For a prompt $x$, the policy $\pi_\theta(\cdot\mid x)$ samples $G$ rollouts $\{y_i\}_{i=1}^{G}$, where $y_i$ denotes the $i$-th rollout response and $A_i$ denotes its assigned advantage; this convention is reused below. Following standard practice in RLVR~\citep{reinforce1,peng2025simko}, REINFORCE in this binary-reward setting fixes the baseline at $b=0.5$ and rescales the reward, so that each rollout contributes a fixed-magnitude signal:
\begin{equation}
\label{eq:rein_adv}
A^{\mathrm{REIN}}_i
=
2\,r(x,y_i)-1
\in
\{+1,-1\}.
\end{equation}
This signal is independent of how well other rollouts for the same prompt performed, making it a reference point for comparing group-relative estimators.

\paragraph{Group-standardized advantage: the GRPO estimator.}
In contrast to the fixed-magnitude REINFORCE signal, \textsc{GRPO}~\citep{shao2024deepseekmath} constructs a sample-dependent advantage from within-group statistics. By standardizing each reward against the empirical mean and standard deviation of the same prompt-level group, GRPO removes the need for a learned value network while reducing gradient variance and providing a relative quality signal that adapts to prompt difficulty. For each prompt $x$, it samples $G$ rollouts $\{y_i\}_{i=1}^{G}$ and computes
\begin{equation}
\label{eq:grpoadv}
A_{i}^{\mathrm{GRPO}}
=
\frac{
r(x,y_i)-\mean\!\big(\{r(x,y_j)\}_{j=1}^{G}\big)
}{
\std\!\big(\{r(x,y_j)\}_{j=1}^{G}\big)
}.
\end{equation}
We refer to $A_{i}^{\mathrm{GRPO}}$ as the \emph{group-relative advantage}, as it evaluates the quality of each rollout relative to its peers within the same prompt-level group.

\paragraph{An accuracy-driven closed form of GRPO.}
Under binary verifiable rewards, the group statistics in Eq.~\eqref{eq:grpoadv} admit a closed form that exposes how the advantage depends on the prompt-level success rate. The mean equals the in-group accuracy
$p=\frac{1}{G}\sum_{j=1}^{G} r(x,y_j)$, and the standard deviation becomes
$\sqrt{p(1-p)}$. Consequently, the advantages for positive ($r=1$) and negative
($r=0$) rollouts in Eq.~\eqref{eq:grpoadv} can be expressed solely as functions
of $p$:
\begin{equation}
\label{adv:grpo}
A_{\mathrm{pos}}^{\mathrm{GRPO}}(p)
=
\sqrt{\frac{1-p}{p}},
\qquad
A_{\mathrm{neg}}^{\mathrm{GRPO}}(p)
=
-\sqrt{\frac{p}{1-p}}.
\end{equation}
Although Eq.~\eqref{adv:grpo} is undefined at $p\in\{0,1\}$, these singularities are vacuous: at $p=0$, no positive rollouts are present, and at $p=1$, no negative rollouts are present. This closed form shows that GRPO weights successes and failures asymmetrically as a function of $p$, motivating a parametric study of modulation strength.

\paragraph{A $\beta$-parametric family unifying REINFORCE and GRPO.}
The REINFORCE form in Eq.~\eqref{eq:rein_adv} and the GRPO form in Eq.~\eqref{adv:grpo} are fixed instantiations: one removes group-accuracy modulation, while the other fixes it to the GRPO scaling. To continuously interpolate between these two fixed instantiations, we introduce a $\beta$-parametrized family that subsumes two channels of advantages:
\begin{equation}
\label{adveq}
A_{\mathrm{pos}}^{(\beta)}(p)
=
\left(\frac{1-p}{p}\right)^{\beta},
\qquad
A_{\mathrm{neg}}^{(\beta)}(p)
=
-\left(\frac{p}{1-p}\right)^{\beta}.
\end{equation}

\begin{wrapfigure}{r}{0.6\textwidth}
\vspace{-0.2em}
    \centering
    \begin{subfigure}[t]{0.48\linewidth}
        \centering
        \includegraphics[width=\linewidth]{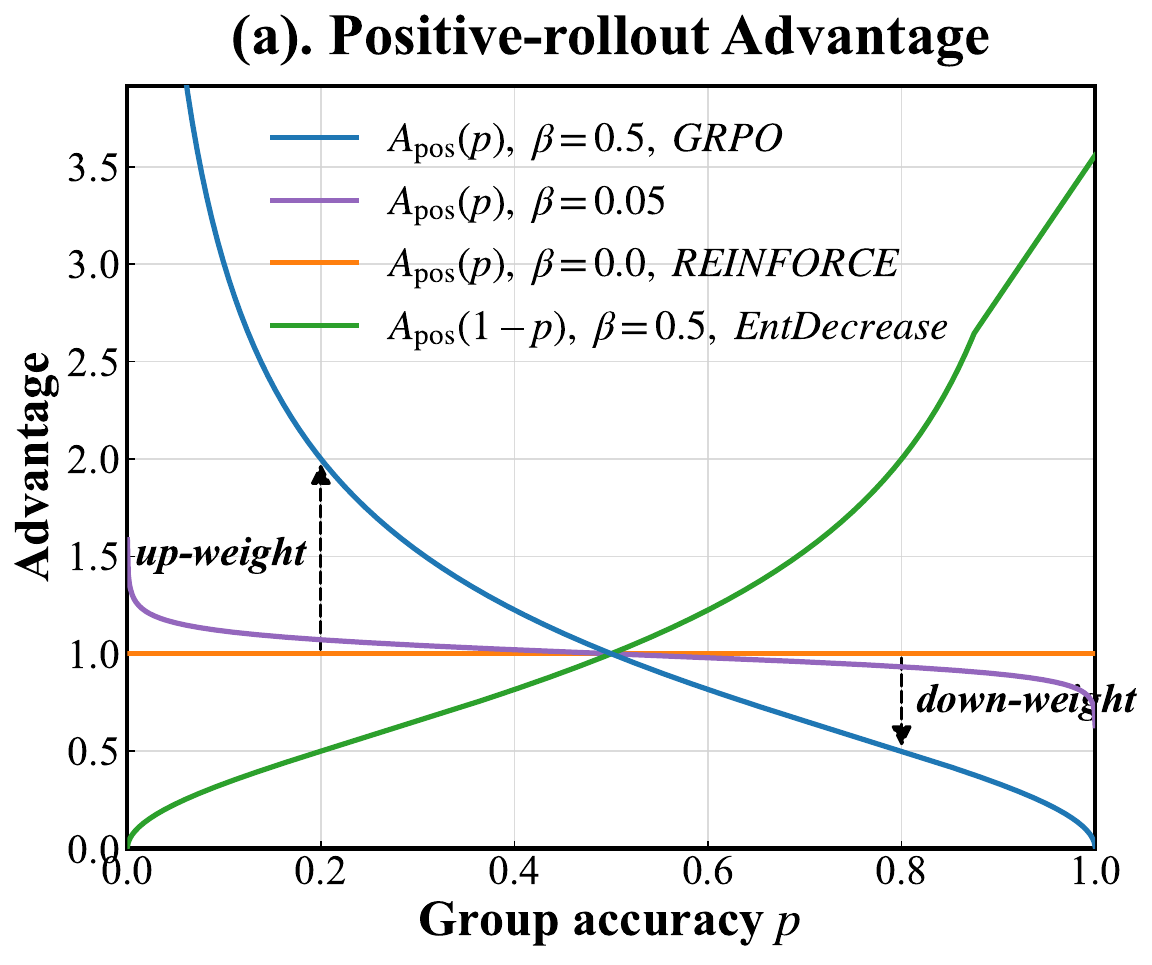}
        \caption*{}
        \label{fig:pos_adv}
    \end{subfigure}
    \hfill
    \begin{subfigure}[t]{0.48\linewidth}
        \centering
        \includegraphics[width=\linewidth]{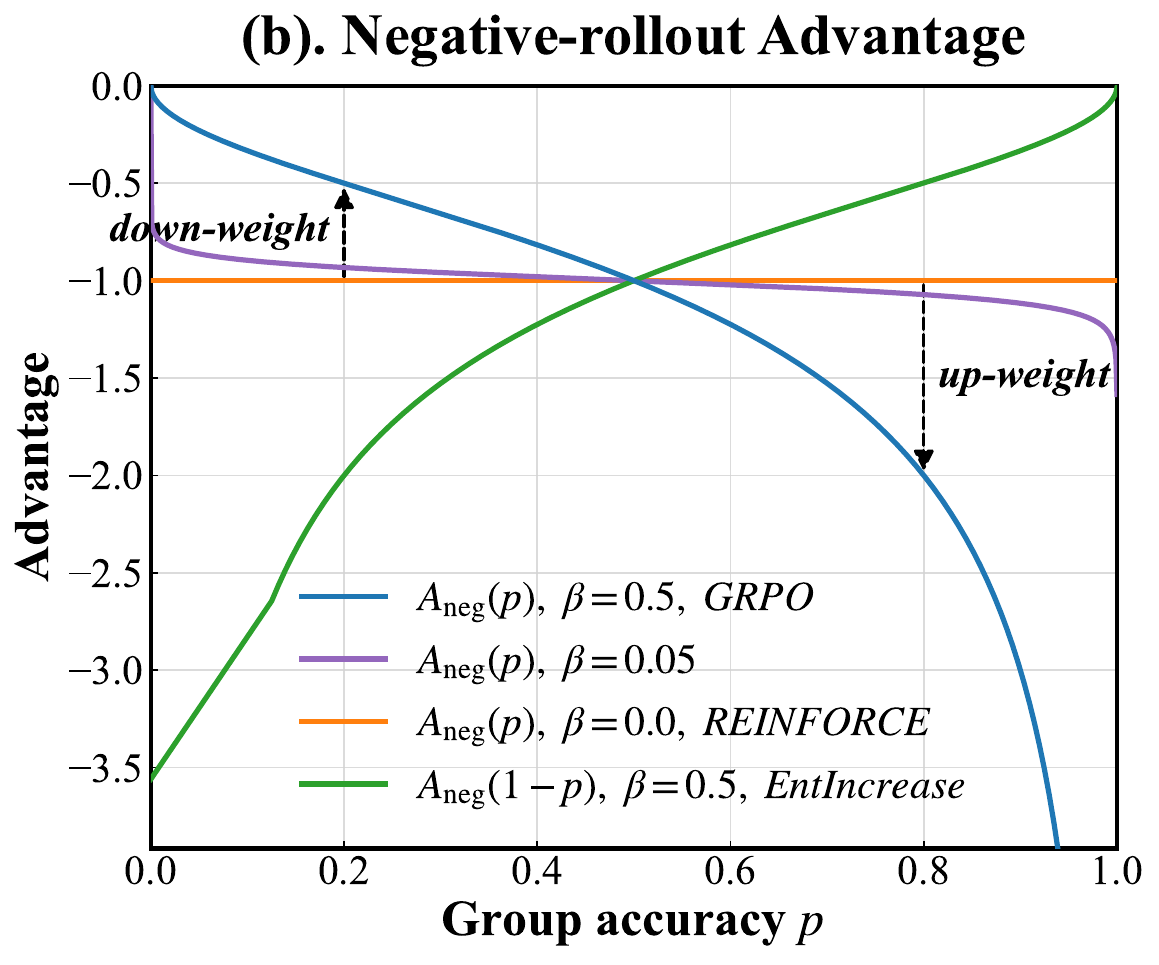}
        \caption*{}
        \label{fig:neg_adv}
    \end{subfigure}
    \vspace{-2.0em}
    \caption{Group-relative advantage magnitudes as functions of group accuracy under the parametric family in Eq.~\eqref{adveq}: (a) successful rollouts and (b) failed rollouts.}
    \label{fig:adv_curve}
    \vspace{-1.2em}
\end{wrapfigure}

Figure~\ref{fig:adv_curve} visualizes advantage magnitudes for successful rollouts in panel (a) and failed rollouts in panel (b) as functions of group accuracy $p$ for representative $\beta$ values. Small values such as $\beta=0.05$ produce mild, near-uniform deviations from REINFORCE, whereas $\beta=0.5$ yields full GRPO modulation.

Blue line in Panel (a) shows that positive rollouts on hard prompts with low $p$ receive amplified advantages, emphasizing rare successful trajectories; blue line in Panel (b) shows that negative rollouts on easy prompts with high $p$ receive amplified penalties, suppressing residual failures. This highlights the difficulty-aware nature of the modulation. Section~2.2 uses this family as the analytical probe for deriving local entropy dynamics.

\subsection{Opposite Entropy Pressures from Positive and Negative Advantage Channels}
\label{subsec:oppo_entropy}
Group-relative advantage is typically viewed as a variance-reduction mechanism in policy gradient estimation for stabilizing RLVR training. We propose to examine it from an entropy perspective because restricted exploration (i.e., fundamental limitation) in RLVR is often reflected by entropy collapse~\citep{cuientropy,yu2025dapo,yue2025does}, and policy entropy directly measures the solution diversity. Characterizing the entropy effect of group-relative advantage is therefore necessary for understanding why GRPO can outperform baselines that share the same policy-gradient objective but differ in advantage construction. In standard GRPO, successful and failed rollouts are modulated simultaneously, with \((\beta_{\mathrm{pos}},\beta_{\mathrm{neg}})=(0.5,0.5)\), so the entropy effects of the two reward channels are coupled. To isolate these effects, we analyze one-sided perturbations from the REINFORCE point \((0,0)\): the positive-channel perturbation \((\beta_+,0)\) applies group-relative modulation only to successful rollouts while keeping failed rollouts at the REINFORCE magnitude, and the negative-channel perturbation \((0,\beta_-)\) does the converse.


\paragraph{Proposition 1 (Proof in Appendix~\ref{app:entropy_response}): Positive- and negative-channel advantage modulation have opposite local effects on entropy.}
\label{prop:opposite_entropy_tendency}
Under the first-order approximation, perturbing only the positive
channel and perturbing only the negative channel around the REINFORCE point
\((\beta_{\mathrm{pos}},\beta_{\mathrm{neg}})=(0,0)\) shift the per-step
entropy change in opposite directions relative to the REINFORCE reference:
\begin{equation}
\label{eq:opposite_entropy_shift}
\Delta H_{\mathcal B}^{(+)}
-
\Delta H_{\mathcal B}^{\mathrm{REIN}}
=
\eta\beta_+\kappa_{\mathcal B}
+
o(\beta_+),
\qquad
\Delta H_{\mathcal B}^{(-)}
-
\Delta H_{\mathcal B}^{\mathrm{REIN}}
=
-\eta\beta_-\kappa_{\mathcal B}
+
o(\beta_-).
\end{equation}
Here, \(\mathcal B\) denotes the prompt batch of one optimization step.
\(\Delta H_{\mathcal B}^{\mathrm{REIN}}\) denotes the entropy change under
\(A^{\mathrm{REIN}}\), while \(\Delta H_{\mathcal B}^{(+)}\) and
\(\Delta H_{\mathcal B}^{(-)}\) denote the corresponding entropy changes under
positive- and negative-channel perturbations, respectively; \(\eta>0\) is the
update step size; and \(\kappa_{\mathcal B}\) is a batch-level
reward-confidence alignment coefficient (full definition in Appendix~\ref{app:entropy_response}). 
For matched perturbation size \(|\beta|\), \textit{\uline{the two leading-order terms in Eq.~\eqref{eq:opposite_entropy_shift} have magnitude \(\eta|\beta|\,|\kappa_{\mathcal B}|\) but opposite signs, so for any prompt batch with the same \(\kappa_{\mathcal B}\), increasing \(\beta_+\) and increasing \(\beta_-\) push the per-step policy-entropy change in opposite directions relative to the REINFORCE reference.}} 
Operationally, group-relative modulation on successful rollouts and on failed rollouts act as separate, sign-coupled forces, and the standard GRPO setting \((\beta_{\mathrm{pos}},\beta_{\mathrm{neg}})=(0.5,0.5)\) activates both at once. The sign of \(\kappa_{\mathcal B}\) determines which channel raises and which channel lowers entropy, while the magnitude of \(\kappa_{\mathcal B}\) scales the strength of this bidirectional effect.

\begin{wraptable}{r}{0.30\textwidth}
\vspace{-1.2em}
\centering
\caption{Estimated \(\kappa_{\mathcal B}\) over the first 20 training steps (scaled by \(10^3\)).}
\label{tab:kappa_early}
\small
\setlength{\tabcolsep}{5pt}
\begin{tabular}{lc}
\toprule
\textbf{Model} & \(\kappa_{\mathcal B}\) \\
\midrule
Qwen2.5-Math-1.5B & \(16.26\) \\
Qwen3-4B          & \(1.91\)  \\
Qwen2.5-7B-Base   & \(6.95\)  \\
\bottomrule
\end{tabular}
\vspace{-1.0em}
\end{wraptable}

We estimate \(\kappa_{\mathcal B}\) over the first 20 training steps
(details in Appendix~\ref{app:kappa_monitoring}) on
Qwen2.5-Math-1.5B~\citep{yang2024qwen2}, Qwen3-4B~\citep{yang2025qwen3}, and
Qwen2.5-7B-Base~\citep{qwen2025qwen25technicalreport}, focusing on the early
training phase where entropy dynamics are most pronounced before the policy
becomes near-deterministic. As shown in Table~\ref{tab:kappa_early},
\(\kappa_{\mathcal B}\) is consistently positive across all evaluated model
backbones. Combined with Proposition~\hyperref[prop:opposite_entropy_tendency]{1},
this indicates that positive-channel modulation contributes
a positive leading-order deviation in per-step entropy change relative to
REINFORCE, while negative-channel modulation contributes a negative one. Thus,
the empirical estimates provide the sign condition needed to interpret the two
channels as local entropy-sustaining and entropy-pruning pressures,
respectively. In Section~\ref{sec:entinfospur}, we empirically verify whether
these locally predicted tendencies persist during training and whether each is
beneficial or harmful for reasoning accuracy.

\section{Linking Channel-Dependent Entropy Changes to Reasoning Accuracy}
\label{sec:entinfospur}
Section~\ref{sec:entropy_mechanism} establishes that the parametric advantage family exerts opposite pressures on policy entropy through the positive and negative reward channels. Two questions remain: whether these locally predicted entropy
tendencies manifest during actual RL training, and whether the resulting entropy changes help or harm reasoning accuracy. To answer these questions, we conduct experiments using Qwen2.5-Math-1.5B~\citep{yang2024qwen2} and Qwen3-4B~\citep{yang2025qwen3}, tracking policy entropy and model test accuracy. Section~3.1 links channel-dependent entropy changes to reasoning accuracy, while Section~3.2 shows that flipping either channel's group-accuracy dependence breaks this benefit. Full settings are provided in Appendix~\ref{app:entropy_exp_setup}.


\subsection{Identifying Productive and Noisy Entropy Through Channel Isolation}
\label{sec:exp_disentangle}
Following the strategy above, we instantiate the parametric family in Eq.~\eqref{adveq} with the four configurations. In terms of the channel-wise modulation strengths \((\beta_{\mathrm{pos}},\beta_{\mathrm{neg}})\), these configurations include the REINFORCE reference \((0,0)\), REINFORCE with entropy regularization, and the Pos-Only and Neg-Only variants, which activate one channel with coefficient \(0.5\) while keeping the other at zero:
\begin{center}
\vspace{-0.3em}
\small
\setlength{\tabcolsep}{6pt}
\begin{tabular}{lll}
\toprule
\textbf{Configuration} & \textbf{Coefficients} & \textbf{Role} \\
\midrule
REINFORCE & \((0,0)\) & constant-magnitude reference \\
Pos-Only & \((0.5,0)\) & modulate successful rollouts only \\
Neg-Only & \((0,0.5)\) & modulate failed rollouts only \\
REINFORCE w/ Ent. Reg. & \((0,0)\) + entropy bonus & uniform entropy-inflation baseline \\
\bottomrule
\end{tabular}
\vspace{-0.5em}
\end{center}
We include REINFORCE with entropy regularization to test whether the entropy
maintained under Pos-Only modulation differs from entropy added uniformly
through a bonus term. We use entropy coefficients calibrated from preliminary
runs, \(\lambda=0.002\) for Qwen2.5-Math-1.5B and \(\lambda=0.005\) for
Qwen3-4B, chosen to make the entropy-regularized baseline visibly increase
entropy and provide a reasonable accuracy gain over the REINFORCE reference
without destabilizing training. The entropy of the current policy
\(\pi_\theta\) over the vocabulary \(\mathcal{V}\) at token position \(t\) is:
\begin{equation}
\label{equ:ent}
\mathcal{H}_t(\pi_\theta)
=
-\sum_{v\in\mathcal{V}}
\pi_\theta(v\mid x,y_{<t})
\log \pi_\theta(v\mid x,y_{<t}).
\end{equation}
\begin{figure*}[t]
  \centering
  \begin{subfigure}[b]{1.0\linewidth}
    \centering
    \includegraphics[width=\linewidth]{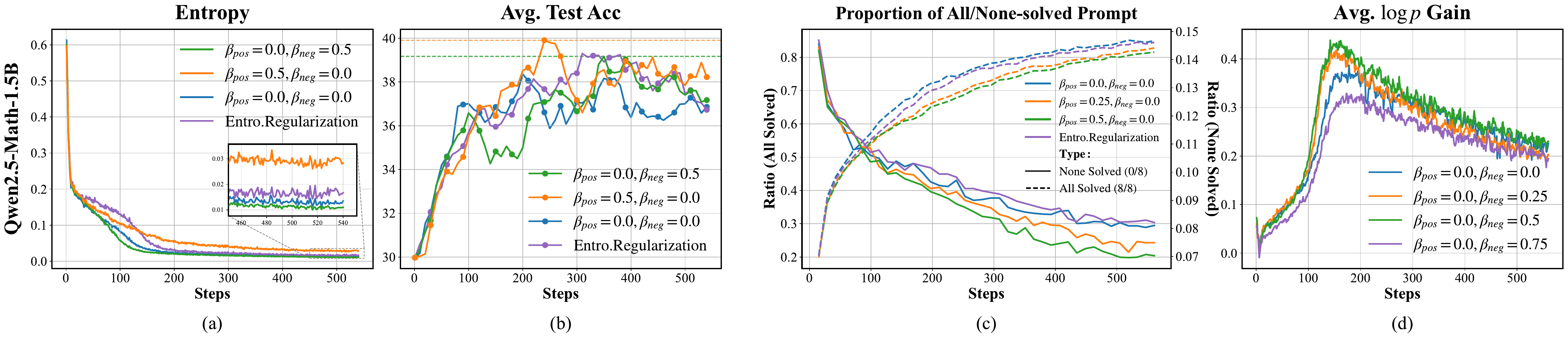}
    \label{fig:top}
    \vspace{-10pt}
  \end{subfigure}
  \vspace{-15pt}
  \begin{subfigure}[b]{1.0\linewidth}
    \centering
    \includegraphics[width=\linewidth]{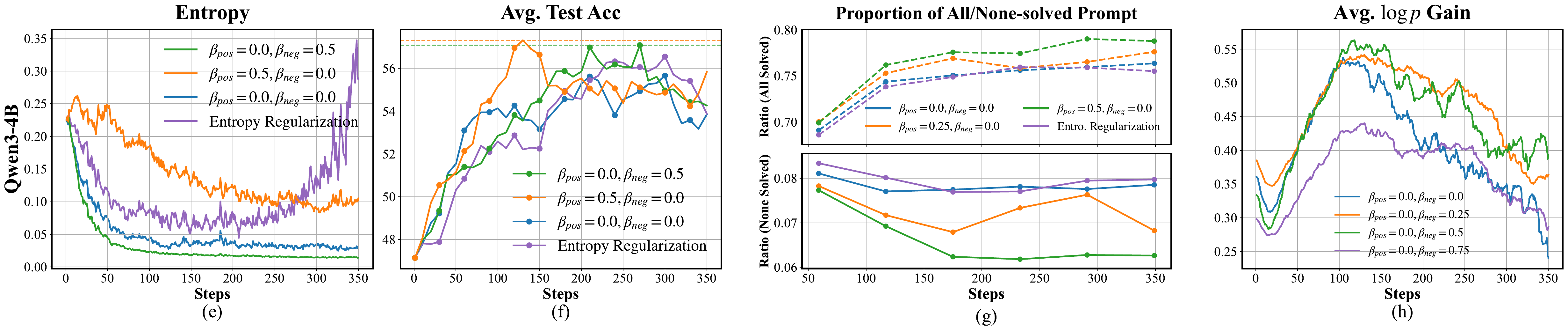}
    \label{fig:bottom}
  \end{subfigure}
  \caption{
        \textbf{Training dynamics and channel-wise diagnostics.}
        The top row reports results on \textbf{Qwen2.5-Math-1.5B}, while the bottom row
        reports results on \textbf{Qwen3-4B}.
        \textbf{(a, e)} Policy entropy over training steps.
        \textbf{(b, f)} Average test accuracy.
        \textbf{(c, g)} Proportions of prompt groups categorized as ``all-solved'' and
        ``none-solved''.
        \textbf{(d, h)} Average log-probability gain of successful rollouts after each
        policy update.
        }
  \vspace{-1em}
  \label{fig:entropyexp1}
\end{figure*}

\noindent\textbf{Pos-Only advantage modulation.}
Under Pos-Only modulation, the policy entropy curve remains above the REINFORCE curve at every training step (Fig.~\ref{fig:entropyexp1}(a,e)), in agreement with Proposition~\hyperref[prop:opposite_entropy_tendency]{1}. This higher entropy is paired with higher test accuracy (Fig.~\ref{fig:entropyexp1}(b,f)), so the additional uncertainty carries usable information for reasoning. We also conduct group analysis: A group is ``none-solved'' if all \(G\) rollouts for a prompt are incorrect, and ``all-solved'' if all \(G\) rollouts are correct. Increasing \(\beta_{\mathrm{pos}}\) monotonically reduces the fraction of none-solved groups (Fig.~\ref{fig:entropyexp1}(c,g)), meaning that the policy starts producing correct rollouts where REINFORCE produced none. Qwen3-4B sees more all-solved groups, while Qwen2.5-Math-1.5B sees fewer; for the latter, this is consistent with reduced over-commitment to easy prompts and with prior reports that over-reinforcing easy instances degrades generalization to harder tasks~\citep{posmech,yao2025debate,dong2025rlpluscounteringcapabilityboundary}. Because the additional entropy under Pos-Only modulation correlates with higher accuracy and a smaller none-solved set, we call it \textbf{productive entropy}.


\noindent\textbf{Neg-Only advantage modulation.}
Under Neg-Only modulation, the policy entropy curve sits below the REINFORCE curve throughout training, with the gap most visible on Qwen3-4B (Fig.~\ref{fig:entropyexp1}(a,e)). This matches Proposition~\hyperref[prop:opposite_entropy_tendency]{1} for the negative channel, and test accuracy nonetheless improves over REINFORCE (Fig.~\ref{fig:entropyexp1}(b,f)), so the part of removed entropy was not contributing to reasoning quality. To understand why removal helps, we track \(\mathbb{E}[\log \pi_{\mathrm{new}}(y)-\log \pi_{\mathrm{old}}(y)]\) over correct rollouts, the average per-step log-probability gain on successes. Increasing \(\beta_{\mathrm{neg}}\) from \(0\) to \(0.5\) raises this gain monotonically (Fig.~\ref{fig:entropyexp1}(d,h)), meaning the policy becomes more confident on its correct trajectories after each update. This is consistent with \textit{Lazy Likelihood Displacement}~\citep{negmech}: because incorrect rollouts often share long reasoning prefixes with correct ones, uniform penalties on failures inadvertently suppress the probability mass of correct and valid paths~\citep{negmech2,negmech3}; group-relative advantage scaling of the negative channel attenuates this interference. The benefit diappears at \(\beta_{\mathrm{neg}}=0.75\), the log-probability gain falls below the REINFORCE baseline, indicating that excessively weak penalties on common failure modes leave the policy unable to escape error patterns on hard prompts~\citep{reinforce1}. Because the part of removed entropy by the negative channel has no measurable benefit and, when modulated correctly, reduces interference on correct paths, we call it \textbf{noisy entropy}.

\noindent\textbf{Entropy regularization.}
Uniform entropy regularization raises the policy entropy curve above REINFORCE, confirming that the bonus performs its intended function. Despite this increase, test accuracy remains below Pos-Only modulation at the best-tuned \(\lambda=0.001\) (Fig.~\ref{fig:entropyexp1}(b,f)). The none-solved and all-solved fractions remain close to REINFORCE (Fig.~\ref{fig:entropyexp1}(c,g)), so added entropy does not translate into solving previously none-solved prompts, unlike Pos-Only modulation. \textit{\uline{Raising entropy uniformly is not equivalent to raising the productive entropy; the reward channel through which the policy entropy is added matters.}}


Together, these results give three complementary observations: the positive channel raises productive entropy and accuracy, the negative channel removes noisy entropy while accuracy still improves, and uniform entropy regularization raises entropy without the corresponding accuracy gain. This pattern is consistent with the \textit{\uline{channel-dependent entropy effect hypothesis}}, in which the channel through which entropy is modulated determines whether the resulting uncertainty is useful. 


\begin{figure*}[t]
  \centering
  \begin{subfigure}[b]{0.49\linewidth}
    \centering
    \includegraphics[width=\linewidth]{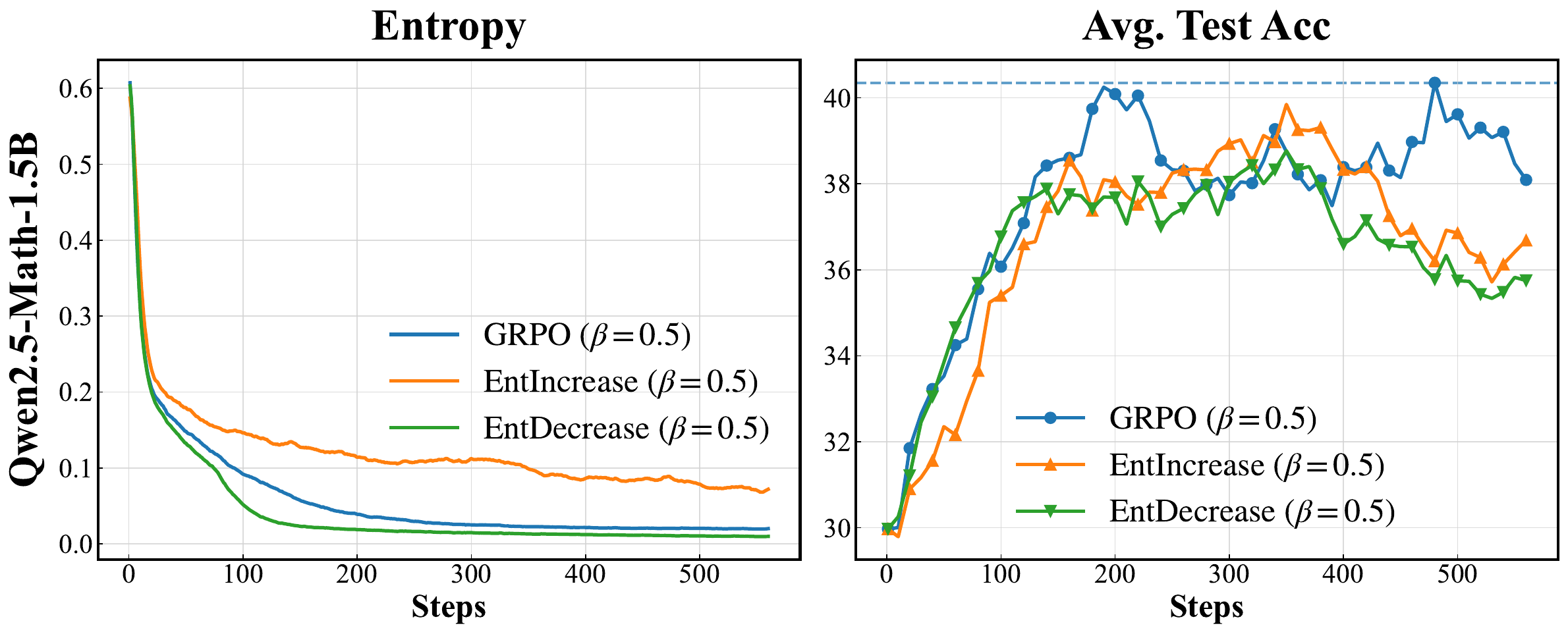}
    \label{fig:sub1}
    \vspace{-1em}
  \end{subfigure}
  \hfill
  \begin{subfigure}[b]{0.49\linewidth}
    \centering
    \includegraphics[width=\linewidth]{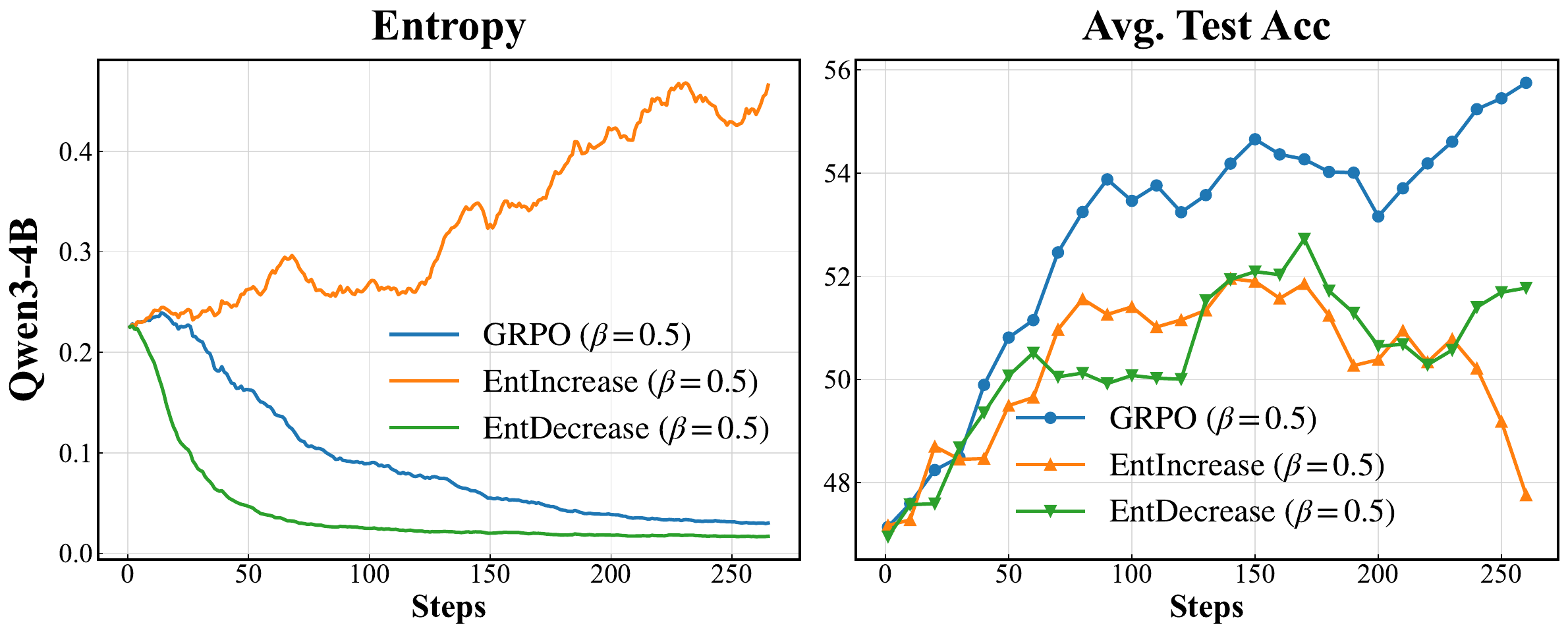}
    \label{fig:sub2}
    \vspace{-1em}
  \end{subfigure}
  \caption{
  \textbf{Advantage Flip Test.}
  \textbf{(a, c)} Policy entropy.
  \textbf{(b, d)} Average test accuracy.
  }
  \vspace{-2em}
  \label{fig:adventropy}
\end{figure*}

\subsection{Advantage Flip Test: The Necessity of Opposite Channel-Wise Entropy Modulation}
\label{sec:exp_adversarial}
The above analysis showed that activating either channel alone improves accuracy, but this does not prove whether the direction of how the positive or negative advantage magnitude changes with group accuracy matters. To rule this out, we fix the modulation strength at standard GRPO, \((\beta_{\mathrm{pos}},\beta_{\mathrm{neg}})=(0.5,0.5)\), and adversarially reverse the accuracy dependence in one channel, asking whether the policy still benefits. Starting from the parametric advantage in Eq.~\eqref{adveq} with \(\beta_{\mathrm{pos}}=\beta_{\mathrm{neg}}=0.5\), we replace \(p\) by \(1-p\) in exactly one channel-dependent function, equivalently reflecting \(\tilde{A}(p)=A(1-p)\) around \(p=0.5\).\footnote{For completeness, advantages at boundary cases are handled by linearly extending the final segment of the curve; full implementation details are provided in Appendix~\ref{app:flipped_curves}.}
The resulting variants are summarized below:
\begin{center}
\vspace{-0.3em}
\footnotesize
\setlength{\tabcolsep}{3pt}
\resizebox{\textwidth}{!}{%
\begin{tabular}{llll}
\toprule
\textbf{Variant} & \textbf{Flipped channel} & \textbf{Closed-form advantage on flipped channel} & \textbf{Entropy tendency} \\
\midrule
GRPO & none & \(A_{\mathrm{pos}}^{(0.5)}(p)=\sqrt{(1-p)/p};\; A_{\mathrm{neg}}^{(0.5)}(p)=-\sqrt{p/(1-p)}\) & original trends \\
EntDecrease & positive flipped; negative unchanged & \(\tilde{A}_{\mathrm{pos}}(p)=\sqrt{p/(1-p)}\) & lower entropy \\
EntIncrease & negative flipped; positive unchanged & \(\tilde{A}_{\mathrm{neg}}(p)=-\sqrt{(1-p)/p}\) & higher entropy \\
\bottomrule
\end{tabular}
}
\vspace{-1em}
\end{center}

\noindent\textbf{Observation.}
EntDecrease makes the policy entropy curve sit below the GRPO curve, while
EntIncrease drives it above (Fig.~\ref{fig:adventropy}(a,c)), confirming that
the flipped curves invert the targeted channel's entropy effect. However, both
variants underperform GRPO and exhibit late-stage degradation in test accuracy
(Fig.~\ref{fig:adventropy}(b,d)). This pattern indicates that the direction of
channel-wise modulation matters: reversing the positive-channel trend removes
variability that is useful under the original GRPO dynamics, whereas reversing
the negative-channel trend injects uncertainty that does not translate into
better reasoning. Together, these adversarial flips support the original GRPO
structure, where positive and negative channels induce complementary entropy
pressures. Reversing either channel disrupts this balance and degrades reasoning performance.

\begin{takeawaybox}
\begin{itemize}[leftmargin=*]
    \item GRPO's positive and negative reward channels affect policy entropy in opposite ways: the positive keeps entropy higher than REINFORCE, while the negative decays entropy faster. 
    \item These two effects have different roles: positive-channel modulation reveals \textbf{productive entropy} tied to improved exploration, while negative-channel modulation removes \textbf{noisy entropy} that interferes with correct trajectories.
    \item Adversarially flipping either channel's group-accuracy dependence hurts test accuracy, showing that the \textbf{direction} of channel-wise modulation matters, not only its magnitude.
    \item \textbf{Conclusion}: Effective RLVR requires \textbf{channel-aware entropy refinement}: selectively sustaining entropy when it supports productive reasoning and pruning entropy when it reflects non-functional uncertainty.
\end{itemize}
\end{takeawaybox}

\section{Asymmetric Group-Relative Policy Optimization}
\label{sec:asymgrpo}
GRPO raises entropy through the positive channel and lowers entropy through the negative channel, and reversing either direction degrades performance. Standard GRPO realizes this mechanism at a single point of the parametric family, \((\beta_{\mathrm{pos}},\beta_{\mathrm{neg}})=(0.5,0.5)\), which forces the modulation strength at the two channels to be tied together. Section~\ref{sec:entinfospur} shows that the two channels play distinct roles: the positive channel expands the solvable set, while the negative channel mitigates Lazy Likelihood Displacement. There is therefore no a prior reason for their optimal modulation strengths to coincide.


We propose \textbf{Asymmetric Group-Relative Policy Optimization (AsymGRPO)}, which decouples \(\beta_{\mathrm{pos}}\) and \(\beta_{\mathrm{neg}}\) as two independent control parameters. For each prompt \(x\sim\mathcal{D}\), the policy samples \(G\) rollouts \(\{y_i\}_{i=1}^{G}\), where \(i\) indexes the rollout, \(y_i=(y_{i,1},\ldots,y_{i,T_i})\) is a token sequence of length \(T_i\), and the token-level advantage \(A_{i,t}\) assigns a scalar learning signal to the \(t\)-th token of rollout \(y_i\). We
define the AsymGRPO token-level advantage as:
\begin{equation}
\label{eq:asym_adv}
\resizebox{\linewidth}{!}{$
\displaystyle
A_{i,t}^{\mathrm{Asym}}(p)
=
\left(\frac{1-p}{p}\right)^{\beta_{\mathrm{pos}}}
\ \text{if } r(x,y_i)=1,
\qquad
A_{i,t}^{\mathrm{Asym}}(p)
=
-\left(\frac{p}{1-p}\right)^{\beta_{\mathrm{neg}}}
\ \text{if } r(x,y_i)=0.
$}
\end{equation}
Setting \(\beta_{\mathrm{pos}}\neq\beta_{\mathrm{neg}}\) activates the asymmetric regime, for example, a high \(\beta_{\mathrm{pos}}\) amplifies advantages on rare correct trajectories while a moderate \(\beta_{\mathrm{neg}}\) penalizes failures without inducing collapse.
As defined in
Eq.~\eqref{eq:asym_adv}, the advantage is computed at the rollout level and
broadcast to all token positions in the response, following standard practice in
RLVR with outcome-level rewards. During training, \(A_{i,t}^{\mathrm{Asym}}\) is plugged into the standard PPO-style clipped objective used by GRPO, changing only the advantage estimator; the full objective is provided in Appendix~\ref{app:ppoloss}.



\subsection{Theoretical Analysis: Understanding the Flexibility of AsymGRPO}
\label{sec:asym_theory}

To understand the flexibility introduced by the decoupled coefficients
\(\beta_{\mathrm{pos}}\) and \(\beta_{\mathrm{neg}}\), we analyze how they affect
the distribution of gradient weights across group-accuracy states. For a group
with accuracy \(p\), the fraction of successful rollouts is \(p\), and the
fraction of failed rollouts is \(1-p\). To focus on the effect of the advantage
estimator, we omit token-level factors, importance ratios, and clipping terms in
this analysis. We denote by
\(\mathcal{W}_{+}(p;\beta_{\mathrm{pos}})\) and
\(\mathcal{W}_{-}(p;\beta_{\mathrm{neg}})\) the total scalar advantage mass
assigned to the positive and negative channels at group-accuracy level \(p\),
respectively:
\begin{equation}
\label{eq:pressure_pos}
\mathcal{W}_{+}(p;\beta_{\mathrm{pos}})
=
p A_{\mathrm{pos}}^{(\beta_{\mathrm{pos}})}(p)
=
p^{1-\beta_{\mathrm{pos}}}(1-p)^{\beta_{\mathrm{pos}}},
\end{equation}
\begin{equation}
\label{eq:pressure_neg}
\mathcal{W}_{-}(p;\beta_{\mathrm{neg}})
=
(1-p)\left|A_{\mathrm{neg}}^{(\beta_{\mathrm{neg}})}(p)\right|
=
p^{\beta_{\mathrm{neg}}}(1-p)^{1-\beta_{\mathrm{neg}}}.
\end{equation}
These quantities summarize the scalar update pressure contributed by each
reward channel at a given group-accuracy level.

\paragraph{Proposition 2 (Proof in Appendix~\ref{app:opt_state_allocation}): Peak Gradient Pressure Across Group Accuracy.}
\label{prop:opt_state_allocation}
For \(\beta_{\mathrm{pos}},\beta_{\mathrm{neg}}\in(0,1)\), the positive and negative channels exert their largest gradient weights at group-accuracy levels determined by their modulation strengths:

\begin{equation}
\label{eq:pressure_peak}
p_{+}^{*}
=
\arg\max_{p\in(0,1)}
\mathcal{W}_{+}(p;\beta_{\mathrm{pos}})
=
1-\beta_{\mathrm{pos}},
\qquad
p_{-}^{*}
=
\arg\max_{p\in(0,1)}
\mathcal{W}_{-}(p;\beta_{\mathrm{neg}})
=
\beta_{\mathrm{neg}}.
\end{equation}

Compared to GRPO, which fixes both peaks at \(p=0.5\), AsymGRPO allows the peak
locations of positive and negative gradient weights to shift independently
along the group-accuracy axis through \(\beta_{\mathrm{pos}}\) and
\(\beta_{\mathrm{neg}}\). This provides additional degrees of freedom for tuning
training dynamics beyond the symmetric GRPO allocation. This controllability
also opens the door to adaptive scheduling strategies that adjust
\(\beta_{\mathrm{pos}}\) and \(\beta_{\mathrm{neg}}\) based on training
dynamics, which we leave to future work.

\begin{table*}[t]
    \centering
    \caption{Main experimental results on mathematical reasoning benchmarks (top: full comparison on Qwen3-4B; bottom: generalization across model backbones). The best result in each column is shown in \textbf{bold}, and the second-best is \underline{underlined}.}
    \label{tab:main_results}
    \small
    \setlength{\tabcolsep}{4pt} 
    \begin{tabular*}{\textwidth}{@{\extracolsep{\fill}}l c c c c c c}
    \toprule
    \textbf{Method} & \textbf{MATH-500} & \textbf{AIME24} & \textbf{AIME25} & \textbf{AMC23} & \textbf{Olympiad} & \textbf{Avg.} \\
    \midrule
    Qwen3-4B & 81.60 & 21.67 & 20.00 & 63.75 & 47.52 & 46.91 \\
    \midrule
    REINFORCE & 86.60 & 28.67 & 24.67 & 73.75 & 54.86 & 53.71 \\
    GRPO~\citep{guo2025deepseek} & 88.20 & 31.00 & 27.33 & 78.25 & 57.74 & 56.50 \\
    GRPO w/ Entro.Regularization & 88.20 & \underline{38.33} & 28.33 & 75.50 & 57.24 & 57.52 \\
    GRPO w/ Clip-higher~\citep{yu2025dapo} & \textbf{90.07} & 34.67 & \textbf{32.33} & \underline{78.50} & \underline{58.18} & \underline{58.75} \\
    Dr.GRPO~\citep{drgrpo} & 88.87 & 36.33 & \underline{30.00} & 78.25 & 57.24 & 58.14 \\
    \midrule
    Pos-Only Modulation (\S~\ref{sec:exp_disentangle}) & 87.13 & 27.33 & 28.00 & 76.75 & 57.34 & 55.31 \\
    Neg-Only Modulation (\S~\ref{sec:exp_disentangle}) & 87.00 & 26.00 & 27.00 & 78.00 & 54.46 & 54.49 \\
    AsymGRPO ($\beta_{\mathrm{pos}}=\beta_{\mathrm{neg}}$) & 88.53 & 32.00 & 29.33 & 78.50 & 57.34 & 57.14 \\
    \rowcolor{gray!15} \textbf{AsymGRPO} & \underline{89.33} & \textbf{39.33} & 28.67 & \textbf{81.00} & \textbf{58.48} & \textbf{59.36} \\
    \bottomrule
    \end{tabular*}
    \vspace{0.8em}
    \label{tab:generalization}
    \small
    \setlength{\tabcolsep}{4pt}
    \begin{tabular*}{\textwidth}{@{\extracolsep{\fill}}l c c c c c c}
    \toprule
    \textbf{Method} & \textbf{MATH-500} & \textbf{AIME24} & \textbf{AIME25} & \textbf{AMC23} & \textbf{Olympiad} & \textbf{Avg.} \\
    \midrule
    Qwen2.5-7B-Base & 44.73 & 5.00 & 0.68 & 25.00 & 21.56 & 19.39 \\
    \midrule
    GRPO & 74.31 & 15.33 & 6.89 & 51.25 & 36.02 & 36.76 \\
    \rowcolor{gray!15} \textbf{AsymGRPO} & \textbf{75.25} & \textbf{16.00} & \textbf{11.03} & \textbf{54.50} & \textbf{37.73} & \textbf{38.90} \\
    \midrule
    Qwen2.5-Math-1.5B & 57.28 & 6.33 & 5.17 & 42.75 & 26.79 & 27.66 \\
    \midrule
    GRPO & 73.24 & 15.33 & 10.69 & 53.00 & 34.93 & 37.44 \\
    \rowcolor{gray!15} \textbf{AsymGRPO} & \textbf{74.45} & \textbf{18.67} & \textbf{13.79} & \textbf{55.25} & \textbf{36.38} & \textbf{39.71} \\
    \bottomrule
    \end{tabular*}
    \vspace{-3pt}
\end{table*}

\subsection{Main Experimental Results and Analysis}
\label{sec:main_results_analysis}

\begin{figure}[t]
  \centering
  \begin{minipage}[c]{0.93\linewidth}
    \centering
    \includegraphics[width=\linewidth]{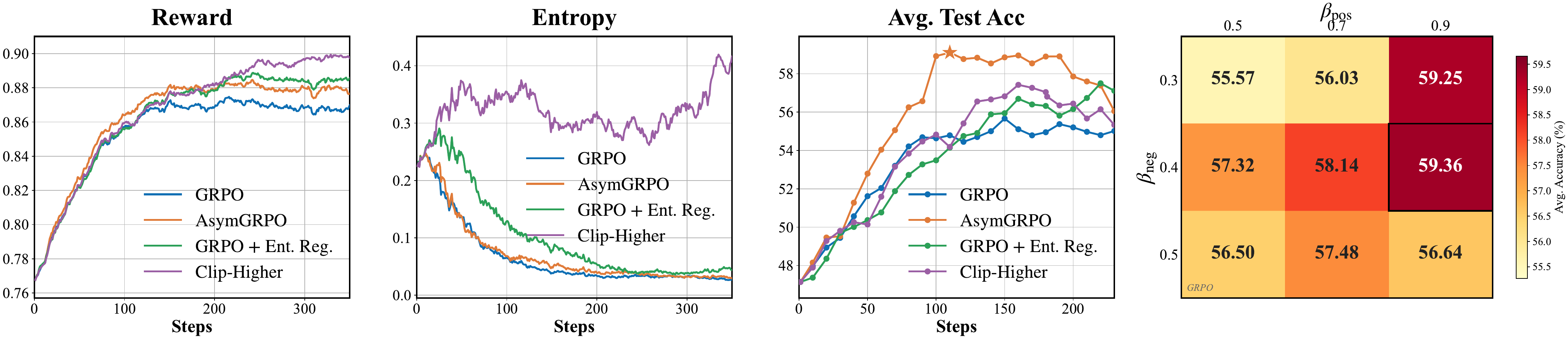}
  \end{minipage}
  \hfill
  
  \caption{
\textbf{Training dynamics and hyperparameter landscape on Qwen3-4B.}
\textbf{(a)} Average training reward.
\textbf{(b)} Policy entropy.
\textbf{(c)} Average test accuracy over training steps.
\textbf{(d)} Average benchmark accuracy across the
\((\beta_{\mathrm{pos}}, \beta_{\mathrm{neg}})\) grid.
All panels use Qwen3-4B as the backbone.
}
\label{fig:main_dynamics}
  \vspace{-10pt}
\end{figure}

\paragraph{Evaluation protocol.}
We evaluate AsymGRPO on mathematical reasoning benchmarks using Qwen3-4B as the
primary backbone, and further test its robustness on Qwen2.5-7B-Base and
Qwen2.5-Math-1.5B. The evaluation suite includes MATH-500, OlympiadBench, AIME
2024, AIME 2025, and AMC 2023. We report Avg@5 accuracy for MATH-500 and
OlympiadBench, and Avg@10 accuracy for AIME 2024, AIME 2025, and AMC 2023, with
temperature \(=0.4\). Table~\ref{tab:main_results} presents the main comparison
and cross-backbone results, while Fig.~\ref{fig:main_dynamics} visualizes the
training dynamics and the \((\beta_{\mathrm{pos}},\beta_{\mathrm{neg}})\) grid. We
summarize the key observations below.

\noindent\textbf{Observation 1: AsymGRPO improves accuracy across baselines and backbones.}
On Qwen3-4B, AsymGRPO achieves the best average accuracy among all listed
methods, reaching \(59.36\%\). It improves over standard GRPO by \(+2.86\)
points (\(56.50\%\rightarrow 59.36\%\)) and over the strongest baseline,
GRPO w/ Clip-higher, by \(+0.61\) points (\(58.75\%\rightarrow 59.36\%\)).
Fig.~\ref{fig:main_dynamics} further shows that this gain does not come from
simply maximizing entropy: AsymGRPO maintains entropy levels comparable to GRPO,
while obtaining stronger reward and test-accuracy dynamics. This suggests
that the improvement comes from better calibration of training pressure through
independent control of the positive and negative channels. The advantage also
generalizes across backbones: AsymGRPO improves over GRPO by \(+2.14\) points on
Qwen2.5-7B-Base (\(36.76\%\rightarrow 38.90\%\)) and by \(+2.27\) points on
Qwen2.5-Math-1.5B (\(37.44\%\rightarrow 39.71\%\)).

\noindent\textbf{Observation 2: Decoupling the two channels is important.}
The symmetric ablation with tied modulation strengths
\((\beta_{\mathrm{pos}}=\beta_{\mathrm{neg}})\) reaches \(57.14\%\), whereas the
decoupled AsymGRPO reaches \(59.36\%\), yielding a \(+2.22\)-point gap. This
indicates that the gain is not merely due to stronger group-relative modulation,
but to the flexibility of controlling the two reward channels separately.
Consistent with Proposition~\hyperref[prop:opt_state_allocation]{2}, decoupling
\(\beta_{\mathrm{pos}}\) and \(\beta_{\mathrm{neg}}\) allows positive and
negative gradient pressure to concentrate on different group-accuracy regions,
which is not possible under a tied symmetric coefficient.

\noindent\textbf{Observation 3: The hyperparameter landscape suggests the need for joint calibration.}
The \((\beta_{\mathrm{pos}},\beta_{\mathrm{neg}})\) grid in
Fig.~\ref{fig:main_dynamics}(d) provides a sensitivity analysis on Qwen3-4B.
Within this grid, stronger positive-channel modulation tends to improve
performance when paired with a moderate negative-channel coefficient, with the
best result achieved at
\((\beta_{\mathrm{pos}},\beta_{\mathrm{neg}})=(0.9,0.4)\).
However, increasing either coefficient is not uniformly beneficial: for example,
\((0.9,0.5)\) drops to \(56.64\%\), close to standard GRPO. This suggests that
the two channels interact and should be calibrated jointly rather than tuned as
independent monotonic knobs.

\noindent\textbf{Observation 4: The two reward channels are complementary.}
The channel-isolated variants improve over the REINFORCE reference:
Pos-Only reaches \(55.31\%\) and Neg-Only reaches \(54.49\%\), compared with
\(53.71\%\) for REINFORCE. However, both remain below GRPO
(\(56.50\%\)), indicating that neither entropy-sustaining positive modulation nor
entropy-pruning negative modulation is sufficient in isolation. Their combination
is necessary for stronger training dynamics, and AsymGRPO further improves on
GRPO by allowing the complementary channels to be controlled independently.

\section{Conclusion}
This work studies entropy dynamics in RLVR through group-relative advantage modulation. We show that GRPO implicitly applies opposite entropy pressures through its positive and negative reward channels: the positive channel sustains productive entropy on successful reasoning trajectories, while the negative channel prunes noisy entropy associated with failed rollouts. This explains why uniform entropy regularization can be insufficient and motivates entropy refinement rather than blind entropy maximization. Building on this view, AsymGRPO decouples the modulation strengths of the two reward channels, enabling flexible control over where gradient pressure concentrates across group-accuracy levels. Experiments on mathematical reasoning benchmarks demonstrate consistent gains over strong RLVR baselines across model backbones. Limitations and future directions are discussed in Appendix~\ref{app:limitations}.





\bibliographystyle{plainnat}
\bibliography{custom}

\appendix
\clearpage

 
        
    
\section{Related Work}


\subsection{Reinforcement Learning for LLM Reasoning}
\label{app:related_work}
Recent post-training research has increasingly focused on reinforcing large
language models (LLMs) in domains such as mathematics and programming using
outcome-level verifiable rewards~\citep{chatgpto1,guo2025deepseek,team2025kimi}.
This paradigm, often referred to as reinforcement learning with verifiable
rewards (RLVR), uses automated correctness feedback to improve long-form
reasoning and has been shown to elicit extended Chain-of-Thought
(CoT) behaviors~\citep{wei2022chain,wang2022self}. A representative example is
DeepSeek-R1~\citep{guo2025deepseek}, which demonstrates that RLVR can
substantially improve reasoning ability and induce behaviors such as
self-reflection and branching during training.

In practice, many RLVR methods optimize PPO-style policy-gradient objectives
while relying on value-free advantage estimators to reduce the cost of reward
baseline estimation~\citep{ppo}. GRPO~\citep{shao2024deepseekmath} estimates
advantages by standardizing rewards within a prompt-level rollout group, while
REINFORCE++~\citep{hu2025reinforce++} uses global advantage normalization to
stabilize policy updates. These estimators are typically motivated by variance
reduction and training stability. In contrast, our work revisits
group-relative advantages from the perspective of entropy dynamics, showing
that the positive and negative reward channels induce distinct entropy effects.
This view motivates AsymGRPO, which modifies the advantage estimator itself
while leaving the PPO-style optimization objective unchanged.

\subsection{Entropy Control and Exploration in RLVR}

Exploration in RLVR is closely tied to the entropy dynamics of LLM policies.
While entropy regularization is a standard mechanism for encouraging
stochasticity in conventional RL~\citep{haarnoja2018soft,ppo}, directly
maximizing entropy in LLM-RL is more delicate because generation occurs over a
large vocabulary and long response horizon~\citep{entropycontrol,sirenentropy}.
Recent work has shown that naive entropy regularization can be highly sensitive
to its coefficient, may inject semantically weak uncertainty, and does not
always translate into better reasoning accuracy~\citep{cui2025entropy,yu2025dapo,yue2025does}.

Existing approaches to entropy control in RLVR can be roughly grouped into two
directions. One line of work maintains entropy at a global level, for example
through target entropy constraints or entropy bonuses that prevent premature
policy collapse~\citep{yu2025dapo,cui2025entropy}. Another line studies the
non-uniform value of uncertainty across the generation process, emphasizing
that reasoning gains often depend on specific high-impact decision points or
``forking'' tokens. Methods in this direction use token pruning or
advantage shaping to concentrate exploration pressure on more informative
parts of the trajectory~\citep{wang2025beyond,entadv,wang2025emergent}.
Concurrent selective-regularization methods further restrict entropy
maximization to more reliable regions, such as top-\(p\) nuclei or
confidence-dependent subsets~\citep{sirenentropy,entropycontrol}.

Our work shares the broad motivation that entropy should be controlled
selectively, but differs in both the object and direction of control. Prior
entropy-control methods primarily seek safer ways to preserve or increase
entropy, for example by restricting entropy bonuses to selected tokens,
confidence regions, or high-impact decision points. In contrast, we show that
effective RLVR training may require a two-sided entropy refinement mechanism:
the positive reward channel sustains productive entropy associated with
successful reasoning trajectories, while the negative reward channel prunes
noisy entropy associated with failed rollouts. This shifts the question from
where to add entropy to how the advantage estimator should allocate both
entropy-sustaining and entropy-pruning pressure. AsymGRPO operationalizes this
view by decoupling the two reward-channel modulation strengths.


\section{Proof of Proposition~\hyperref[prop:opposite_entropy_tendency]{1}}
\label{app:entropy_response}

\begin{proof}
In this section, we prove Proposition~\hyperref[prop:opposite_entropy_tendency]{1}.
We first introduce the notation used in the derivation. For a prompt batch
\(\mathcal B\), let \(g\) index a prompt-level rollout group and \(i\) index a
rollout within the group. Define
\[
Z_{g,i}=\mathbf{1}[r_{g,i}=1],
\qquad
p_g=\frac{1}{G}\sum_{i=1}^{G}Z_{g,i}
=\mathbb{E}_g[Z],
\qquad
L_{g,i}=\log \pi(y_{g,i}\mid x_g),
\]
where \(\mathbb{E}_g[\cdot]\) denotes the empirical average over rollouts in
group \(g\). In the following batch-level expectations, we consider
mixed-outcome groups with \(0<p_g<1\), where both successful and failed rollouts
are present. In practice, \(L_{g,i}\) is computed as the
response-length-normalized rollout log-probability.

For each mixed-outcome group, define the within-group reward-confidence gap,
group-accuracy variance, and reward log-odds as
\[
\delta_g
=
\mathbb{E}_{i:Z_{g,i}=1}[L_{g,i}]
-
\mathbb{E}_{i:Z_{g,i}=0}[L_{g,i}],
\qquad
w_g=p_g(1-p_g),
\qquad
\ell_g=\log\frac{p_g}{1-p_g}.
\]
The aggregate \textbf{reward-confidence alignment coefficient} is
\[
\kappa_{\mathcal B}
=
\mathbb{E}_{g\in\mathcal B}[w_g\delta_g\ell_g]
=
\mathbb{E}_{g\in\mathcal B}
\left[
p_g(1-p_g)\delta_g
\log\frac{p_g}{1-p_g}
\right].
\]

Following \citealp{cui2025entropy}, we use the response-level first-order
entropy-change approximation. For a rollout-level advantage assignment \(A\),
the local entropy change of group \(g\) satisfies
\[
\Delta H_g^{A}
\approx
-\eta\,\operatorname{Cov}_g(L,A),
\]
where \(\eta>0\) is the infinitesimal update step size and
\(\operatorname{Cov}_g\) is the empirical covariance over rollouts in group
\(g\). Hence, for two advantage assignments \(A\) and \(A'\),
\[
\Delta H_g^{A}-\Delta H_g^{A'}
\approx
-\eta\,\operatorname{Cov}_g(L,A-A').
\]

We first record a useful identity. Since
\(\mathbb{E}_g[Z]=p_g\) and
\(\mathbb{E}_g[L]
=
p_g\mathbb{E}[L\mid Z=1]
+
(1-p_g)\mathbb{E}[L\mid Z=0]\),
we have
\[
\operatorname{Cov}_g(L,Z)
=
p_g(1-p_g)
\left(
\mathbb{E}_{i:Z_{g,i}=1}[L_{g,i}]
-
\mathbb{E}_{i:Z_{g,i}=0}[L_{g,i}]
\right)
=
w_g\delta_g.
\]
Consequently,
\[
\operatorname{Cov}_g(L,1-Z)
=
-\operatorname{Cov}_g(L,Z)
=
-w_g\delta_g.
\]

The REINFORCE reference assigns
\(A_{g,i}^{\mathrm{REIN}}=+1\) when \(Z_{g,i}=1\) and
\(A_{g,i}^{\mathrm{REIN}}=-1\) when \(Z_{g,i}=0\). For the positive-channel
perturbation, only successful rollouts are modulated:
\[
A_{g,i}^{(+)}
=
\left(\frac{1-p_g}{p_g}\right)^{\beta_+}
\ \text{if } Z_{g,i}=1,
\qquad
A_{g,i}^{(+)}=-1
\ \text{if } Z_{g,i}=0.
\]
Using \(\ell_g=\log\frac{p_g}{1-p_g}\), this gives
\[
A_{g,i}^{(+)}-A_{g,i}^{\mathrm{REIN}}
=
\left(e^{-\beta_+\ell_g}-1\right)Z_{g,i}.
\]
Therefore,
\[
\begin{aligned}
\Delta H_g^{(+)}-\Delta H_g^{\mathrm{REIN}}
&\approx
-\eta\,\operatorname{Cov}_g
\left(L,A^{(+)}-A^{\mathrm{REIN}}\right) \\
&=
-\eta\left(e^{-\beta_+\ell_g}-1\right)\operatorname{Cov}_g(L,Z) \\
&=
\eta w_g\delta_g\left(1-e^{-\beta_+\ell_g}\right).
\end{aligned}
\]
Averaging over groups and applying the first-order expansion
\(1-e^{-\beta_+\ell_g}=\beta_+\ell_g+o(\beta_+)\), we obtain
\[
\mathbb{E}_{g\in\mathcal B}
\left[
\Delta H_g^{(+)}-\Delta H_g^{\mathrm{REIN}}
\right]
=
\eta\beta_+
\mathbb{E}_{g\in\mathcal B}[w_g\delta_g\ell_g]
+
o(\beta_+).
\]

For the negative-channel perturbation, only failed rollouts are modulated:
\[
A_{g,i}^{(-)}=+1
\ \text{if } Z_{g,i}=1,
\qquad
A_{g,i}^{(-)}
=
-\left(\frac{p_g}{1-p_g}\right)^{\beta_-}
\ \text{if } Z_{g,i}=0.
\]
Thus,
\[
A_{g,i}^{(-)}-A_{g,i}^{\mathrm{REIN}}
=
-\left(e^{\beta_-\ell_g}-1\right)(1-Z_{g,i}).
\]
Using the covariance identity above,
\[
\begin{aligned}
\Delta H_g^{(-)}-\Delta H_g^{\mathrm{REIN}}
&\approx
-\eta\,\operatorname{Cov}_g
\left(L,A^{(-)}-A^{\mathrm{REIN}}\right) \\
&=
\eta\left(e^{\beta_-\ell_g}-1\right)\operatorname{Cov}_g(L,1-Z) \\
&=
-\eta w_g\delta_g\left(e^{\beta_-\ell_g}-1\right).
\end{aligned}
\]
Averaging over groups and using
\(e^{\beta_-\ell_g}-1=\beta_-\ell_g+o(\beta_-)\), we get
\[
\mathbb{E}_{g\in\mathcal B}
\left[
\Delta H_g^{(-)}-\Delta H_g^{\mathrm{REIN}}
\right]
=
-\eta\beta_-
\mathbb{E}_{g\in\mathcal B}[w_g\delta_g\ell_g]
+
o(\beta_-).
\]
Since \(\kappa_{\mathcal B}=\mathbb{E}_{g\in\mathcal B}[w_g\delta_g\ell_g]\),
the two expansions become
\[
\mathbb{E}_{g\in\mathcal B}
\left[
\Delta H_g^{(+)}-\Delta H_g^{\mathrm{REIN}}
\right]
=
\eta\beta_+\kappa_{\mathcal B}+o(\beta_+),
\qquad
\mathbb{E}_{g\in\mathcal B}
\left[
\Delta H_g^{(-)}-\Delta H_g^{\mathrm{REIN}}
\right]
=
-\eta\beta_-\kappa_{\mathcal B}+o(\beta_-).
\]
This proves Proposition~\hyperref[prop:opposite_entropy_tendency]{1}.
\end{proof}

\section{Implementation Details for Advantage-Flip Experiments}
\label{app:flipped_curves}

In Section~\ref{sec:exp_adversarial}, we construct adversarial variants by
reversing the group-accuracy dependence of one reward channel while keeping the
other channel unchanged. For the selected channel, this is implemented by
reflecting the corresponding advantage curve around \(p=0.5\), i.e.,
\[
\tilde A(p)=A(1-p).
\]
This reflection reverses the monotonic weighting trend of the original
group-relative advantage and allows us to test whether the direction of the
channel-wise modulation is important.

For a rollout group of size \(G\), the group accuracy takes discrete values
\(p\in\{0,\frac{1}{G},\ldots,1\}\). The reflected curve introduces one undefined
endpoint for each class-conditional advantage: the flipped positive curve is
undefined at \(p=1\), while the flipped negative curve is undefined at \(p=0\).
We handle these endpoint values by linear extrapolation from the nearest two
feasible group-accuracy points.

Let
\[
B_{\mathrm{pos}}(p)=A_{\mathrm{pos}}^{(\beta)}(1-p),
\qquad
B_{\mathrm{neg}}(p)=A_{\mathrm{neg}}^{(\beta)}(1-p)
\]
denote the reflected positive and negative advantage curves wherever they are
defined. For the positive channel, we extrapolate the endpoint value at \(p=1\)
from the two nearest feasible points:
\[
V_{\mathrm{pos}}^{\mathrm{end}}
=
B_{\mathrm{pos}}\!\left(\frac{G-1}{G}\right)
+
\left[
B_{\mathrm{pos}}\!\left(\frac{G-1}{G}\right)
-
B_{\mathrm{pos}}\!\left(\frac{G-2}{G}\right)
\right].
\]
For the negative channel, we extrapolate the endpoint value at \(p=0\) from the
two nearest feasible points:
\[
V_{\mathrm{neg}}^{\mathrm{end}}
=
B_{\mathrm{neg}}\!\left(\frac{1}{G}\right)
-
\left[
B_{\mathrm{neg}}\!\left(\frac{2}{G}\right)
-
B_{\mathrm{neg}}\!\left(\frac{1}{G}\right)
\right].
\]

We use \(\operatorname{Lin}(p;a,b;u,v)\) to denote linear interpolation between
\((a,u)\) and \((b,v)\):
\[
\operatorname{Lin}(p;a,b;u,v)
=
u+\frac{p-a}{b-a}(v-u).
\]
The final flipped positive advantage is
\[
\tilde{A}_{\mathrm{pos}}(p)
=
\begin{cases}
B_{\mathrm{pos}}(p),
& 0<p\le \frac{G-1}{G}, \\[6pt]
\operatorname{Lin}\!\left(
p;
\frac{G-1}{G},1;
B_{\mathrm{pos}}\!\left(\frac{G-1}{G}\right),
V_{\mathrm{pos}}^{\mathrm{end}}
\right),
& \frac{G-1}{G}<p\le 1,
\end{cases}
\]
and the final flipped negative advantage is
\[
\tilde{A}_{\mathrm{neg}}(p)
=
\begin{cases}
\operatorname{Lin}\!\left(
p;
0,\frac{1}{G};
V_{\mathrm{neg}}^{\mathrm{end}},
B_{\mathrm{neg}}\!\left(\frac{1}{G}\right)
\right),
& 0\le p<\frac{1}{G}, \\[6pt]
B_{\mathrm{neg}}(p),
& \frac{1}{G}\le p<1.
\end{cases}
\]
In our advantage-flip experiments, \(\tilde{A}_{\mathrm{pos}}\) is used for
EntDecrease and \(\tilde{A}_{\mathrm{neg}}\) is used for EntIncrease, as
visualized in Fig.~\ref{fig:adv_curve}(a) and Fig.~\ref{fig:adv_curve}(b),
respectively.

\section{Proof of Proposition~\hyperref[prop:opt_state_allocation]{2}}
\label{app:opt_state_allocation}

\begin{proof}
In this section, we prove Proposition~\hyperref[prop:opt_state_allocation]{2}.
For \(p\in(0,1)\) and \(\beta_{\mathrm{pos}},\beta_{\mathrm{neg}}\in(0,1)\),
the positive and negative scalar learning-pressure functions are
\[
\mathcal{W}_{+}(p;\beta_{\mathrm{pos}})
=
pA_{\mathrm{pos}}^{(\beta_{\mathrm{pos}})}(p)
=
p^{1-\beta_{\mathrm{pos}}}(1-p)^{\beta_{\mathrm{pos}}},
\]
and
\[
\mathcal{W}_{-}(p;\beta_{\mathrm{neg}})
=
(1-p)\left|A_{\mathrm{neg}}^{(\beta_{\mathrm{neg}})}(p)\right|
=
p^{\beta_{\mathrm{neg}}}(1-p)^{1-\beta_{\mathrm{neg}}}.
\]
Both functions are strictly positive on \((0,1)\), so maximizing each function
is equivalent to maximizing its logarithm.

For the positive channel,
\[
\log \mathcal{W}_{+}(p;\beta_{\mathrm{pos}})
=
(1-\beta_{\mathrm{pos}})\log p
+
\beta_{\mathrm{pos}}\log(1-p).
\]
Taking the derivative with respect to \(p\) gives
\[
\frac{\partial}{\partial p}
\log \mathcal{W}_{+}(p;\beta_{\mathrm{pos}})
=
\frac{1-\beta_{\mathrm{pos}}}{p}
-
\frac{\beta_{\mathrm{pos}}}{1-p}.
\]
The unique stationary point satisfies
\[
\frac{1-\beta_{\mathrm{pos}}}{p}
=
\frac{\beta_{\mathrm{pos}}}{1-p},
\]
which yields \(p=1-\beta_{\mathrm{pos}}\). Moreover,
\[
\frac{\partial^2}{\partial p^2}
\log \mathcal{W}_{+}(p;\beta_{\mathrm{pos}})
=
-\frac{1-\beta_{\mathrm{pos}}}{p^2}
-
\frac{\beta_{\mathrm{pos}}}{(1-p)^2}
<0,
\]
so \(\log \mathcal{W}_{+}\), and hence \(\mathcal{W}_{+}\), is strictly concave
on \((0,1)\). Therefore,
\[
p_{+}^{*}
=
\arg\max_{p\in(0,1)}
\mathcal{W}_{+}(p;\beta_{\mathrm{pos}})
=
1-\beta_{\mathrm{pos}}.
\]

The negative channel follows symmetrically. We have
\[
\log \mathcal{W}_{-}(p;\beta_{\mathrm{neg}})
=
\beta_{\mathrm{neg}}\log p
+
(1-\beta_{\mathrm{neg}})\log(1-p),
\]
and therefore
\[
\frac{\partial}{\partial p}
\log \mathcal{W}_{-}(p;\beta_{\mathrm{neg}})
=
\frac{\beta_{\mathrm{neg}}}{p}
-
\frac{1-\beta_{\mathrm{neg}}}{1-p}.
\]
Setting the derivative to zero gives \(p=\beta_{\mathrm{neg}}\). The second
derivative is
\[
\frac{\partial^2}{\partial p^2}
\log \mathcal{W}_{-}(p;\beta_{\mathrm{neg}})
=
-\frac{\beta_{\mathrm{neg}}}{p^2}
-
\frac{1-\beta_{\mathrm{neg}}}{(1-p)^2}
<0,
\]
so this stationary point is the unique global maximizer:
\[
p_{-}^{*}
=
\arg\max_{p\in(0,1)}
\mathcal{W}_{-}(p;\beta_{\mathrm{neg}})
=
\beta_{\mathrm{neg}}.
\]
This proves Proposition~\hyperref[prop:opt_state_allocation]{2}.
\end{proof}

\section{Online Estimation of \texorpdfstring{\(\kappa_{\mathcal B}\)}{kappa}}
\label{app:kappa_monitoring}

We estimate \(\kappa_{\mathcal B}\) online from rollout batches to diagnose
early-stage reward-confidence alignment. The statistic is computed after rollout
rewards are assigned and before any policy update is applied, so it reflects the
policy that generated the responses. It is recorded once per rollout batch and is
not recomputed for repeated PPO minibatches or update epochs.

For each prompt group \(g\) with \(G\) sampled responses, we compute the
response-length-normalized old-policy log-probability
\[
L_{g,i}
=
\frac{
\sum_t m_{g,i,t}\log \pi_{\mathrm{old}}(y_{g,i,t}\mid x_g,y_{g,i,<t})
}{
\sum_t m_{g,i,t}
},
\]
where \(m_{g,i,t}\) is the response mask. The stored old-policy values are
already log-probabilities, so no additional logarithm is applied. Let
\(r_{g,i}\in\{0,1\}\) be the binary reward and
\(p_g=\frac{1}{G}\sum_{i=1}^{G}r_{g,i}\) be the group accuracy. We exclude
groups with \(p_g=0\) or \(p_g=1\), since they contain only one reward class and
therefore have no within-group reward-confidence contrast.

For each mixed-outcome group, we compute the confidence gap
\(\delta_g=\mathbb{E}[L_{g,i}\mid r_{g,i}=1]
-\mathbb{E}[L_{g,i}\mid r_{g,i}=0]\), the log-odds term
\(\ell_g=\log\frac{p_g}{1-p_g}\), and the mixedness weight
\(w_g=p_g(1-p_g)\). The group-level contribution is
\(k_g=w_g\delta_g\ell_g\). For a rollout batch \(\mathcal B\), we estimate
\[
\kappa_{\mathcal B}
=
\frac{1}{|\mathcal V_{\mathcal B}|}
\sum_{g\in\mathcal V_{\mathcal B}} k_g,
\]
where \(\mathcal V_{\mathcal B}\) denotes the set of mixed-outcome groups. This
inner group-level average matches the definition of \(\kappa_{\mathcal B}\) in
Proposition~\hyperref[prop:opposite_entropy_tendency]{1}

To summarize early training behavior, we average the step-level estimates
\[
\widehat{\kappa}
=
\frac{1}{T}
\sum_{s=1}^{T}
\kappa_{\mathcal B_s}.
\]
This step-level average matches the per-step interpretation of the local entropy
analysis. We focus on the early stage because
Proposition~\hyperref[prop:opposite_entropy_tendency]{1} characterizes a local tendency
near the initial REINFORCE-like regime, and entropy changes are most pronounced
before the policy becomes substantially more deterministic.

\section{PPO-style Optimization with Decoupled Advantages}
\label{app:ppoloss}
To make AsymGRPO a drop-in replacement for GRPO, we keep the standard PPO-style clipped surrogate objective at the token level and modify only the advantage estimator. For each prompt \(x\), we sample rollouts \(\{y_i\}_{i=1}^{G}\sim \pi_{\theta_{\mathrm{old}}}(\cdot\mid x)\) and update \(\theta\) by maximizing the AsymGRPO objective:

\begin{equation}
\label{eq:asym_ppo}
\resizebox{\linewidth}{!}{$
\displaystyle
\mathcal{J}_{\mathrm{Asym}}(\theta)
=
\mathbb{E}_{x\sim\mathcal{D},\,\{y_i\}_{i=1}^{G}\sim\pi_{\theta_{\mathrm{old}}}}
\left[
\frac{1}{G}\sum_{i=1}^{G}
\frac{1}{T_i}\sum_{t=1}^{T_i}
\min\!\left(
\rho_{i,t}(\theta) A_{i,t}^{\mathrm{Asym}},
\operatorname{clip}\!\left(\rho_{i,t}(\theta),1-\epsilon,1+\epsilon\right)
A_{i,t}^{\mathrm{Asym}}
\right)
\right].
$}
\end{equation}
Here,
\(\rho_{i,t}(\theta)=
\frac{\pi_{\theta}(y_{i,t}\mid x,y_{i,<t})}
{\pi_{\theta_{\mathrm{old}}}(y_{i,t}\mid x,y_{i,<t})}\)
is the token-level importance ratio and \(\epsilon\) is the clipping hyperparameter. 
Thus, AsymGRPO preserves the standard PPO objective and modifies only the
advantage estimator, decoupling the learning signals from successful and failed
rollouts through \(\beta_{\mathrm{pos}}\) and \(\beta_{\mathrm{neg}}\).

\section{Experimental Settings for Entropy Dynamics Analysis (Section~\ref{sec:entinfospur})}
\label{app:entropy_exp_setup}

This section details the experimental setup used to examine the impact of entropy dynamics on reasoning performance. All RL experiments are implemented using the verl~\cite{sheng2025hybridflow} framework on a single node equipped with 4 $\times$ NVIDIA H100 GPUs.

We conduct ablation studies on two base models: Qwen2.5-Math-1.5B~\citep{yang2024qwen2} and Qwen3-4B~\citep{yang2025qwen3}. For Qwen3-4B, we specifically utilize its non-thinking mode for training. The models are trained on the MATH dataset~\cite{hendrycks2021measuring}, which contains 7,500 problems spanning diverse mathematical areas and difficulty levels.

We employ the AdamW optimizer with a learning rate of $2 \times 10^{-6}$ for both models. Following \citealp{yu2025dapo}, we apply token-level loss aggregation for all settings. For each query, the policy generates $G=8$ rollouts. Regarding model-specific configurations, the Qwen2.5-Math-1.5B experiments use a global batch size of 512, a mini-batch size of 128, and a maximum response length of 2,560 tokens. Conversely, the Qwen3-4B experiments utilize a global batch size of 128, a mini-batch size of 64, and a maximum response length of 4,096 tokens.

To monitor performance, we report the Avg. Test Accuracy, calculated as the mean accuracy across five mathematical reasoning benchmarks: AIME 2024, AIME 2025, MATH-500~\cite{math500}, AMC 2023 and OlympiadBench~\cite{he2024olympiadbench}. Evaluation is performed every 10 training steps and the temperature is set to $0$ to ensure fast and reliable evaluation of model capabilities. To clearly visualize training trends, we apply Exponential Moving Average (EMA) smoothing with a factor of $0.7$ to all test accuracy curves.

\section{Experimental Settings for Main Results (Section~\ref{sec:asymgrpo})}
\label{app:main_exp_setup}

\paragraph{Training setup.}
We conduct the main experiments using the verl~\citep{sheng2025hybridflow}
framework on a single node equipped with \(4\times\) NVIDIA H100 GPUs. The
training configuration follows Appendix~\ref{app:entropy_exp_setup}, including
the optimizer, rollout group size, token-level loss aggregation, batch-size
settings, and maximum response length. We use \textbf{Qwen3-4B} as the primary
backbone and train on the MATH training set~\citep{hendrycks2021measuring}. To
evaluate robustness across model backbones, we additionally run experiments on
\textbf{Qwen2.5-7B-Base} and \textbf{Qwen2.5-Math-1.5B}~\citep{yang2024qwen2}.

\paragraph{Evaluation protocol.}
We evaluate all methods on five mathematical reasoning benchmarks:
MATH-500~\citep{math500}, OlympiadBench~\citep{he2024olympiadbench},
AIME 2024, AIME 2025, and AMC 2023. Following the main text, we report
Avg@5 accuracy for MATH-500 and OlympiadBench, and Avg@10 accuracy for
AIME 2024, AIME 2025, and AMC 2023, with temperature \(=0.4\). The final
average score is computed over these five benchmarks. Across methods, we keep
the training and decoding configurations matched for controlled comparison.

\paragraph{Baselines and variants.}
For the Qwen3-4B main comparison, we compare AsymGRPO against standard
GRPO~\citep{guo2025deepseek}, GRPO with entropy regularization, GRPO with
Clip-higher~\citep{yu2025dapo}, and Dr.GRPO~\citep{drgrpo}. We also include
the channel-isolated variants from Section~\ref{sec:exp_disentangle}:
Pos-Only modulation \((\beta_{\mathrm{pos}},\beta_{\mathrm{neg}})=(0.5,0)\)
and Neg-Only modulation \((0,0.5)\), together with a symmetric AsymGRPO
ablation using \(\beta_{\mathrm{pos}}=\beta_{\mathrm{neg}}=0.7\). For entropy
regularization, we use \(\lambda=0.001\). For Clip-higher, we set the upper
clipping threshold to \(\epsilon_{\mathrm{high}}=0.28\). For AsymGRPO, we set
\(\beta_{\mathrm{pos}}=0.9\) and \(\beta_{\mathrm{neg}}=0.4\).

\section{Limitations}
\label{app:limitations}

Although AsymGRPO achieves consistent improvements across mathematical reasoning benchmarks and model backbones, our study has several limitations. First, AsymGRPO currently uses fixed values of \(\beta_{\mathrm{pos}}\) and \(\beta_{\mathrm{neg}}\) throughout training, and these coefficients are selected through hyperparameter tuning. While our hyperparameter grid and theoretical analysis provide heuristic guidance for choosing these coefficients, we have not yet explored adaptive scheduling strategies. Proposition~\hyperref[prop:opt_state_allocation]{2} shows that \(\beta_{\mathrm{pos}}\) and \(\beta_{\mathrm{neg}}\) directly determine where positive and negative gradient pressure concentrates along the group-accuracy axis, providing a useful basis for future training-dynamics-aware schedules. Second, due to resource constraints, our empirical evaluation focuses on mathematical reasoning benchmarks, and we have not yet evaluated AsymGRPO on broader verifiable-reward domains such as code generation or other structured reasoning tasks.

\section{Licenses}
Qwen3~\cite{yang2025qwen3}, Qwen2.5~\cite{qwen2025qwen25technicalreport} and Qwen2.5-Math~\cite{yang2024qwen2} are distributed under the Apache License 2.0. The MATH dataset~\cite{hendrycks2021measuring} and its subset MATH-500~\cite{math500} are released under the MIT license. The OlympiadBench dataset~\cite{he2024olympiadbench} is released under the Creative Commons Attribution-NonCommercial 4.0 (CC BY-NC 4.0) license. The AIME and AMC datasets are utilized strictly for academic research and evaluation purposes. All resources are used in accordance with their respective licensing terms.

\end{document}